\documentclass[11pt]{article}

\usepackage[table]{xcolor}
\usepackage[final]{acl}

\usepackage{times}
\usepackage{latexsym}
\usepackage{amsmath,amssymb}

\usepackage[T1]{fontenc}
\usepackage[utf8]{inputenc}

\usepackage{microtype}

\usepackage{inconsolata}

\usepackage{graphicx}
\usepackage{algorithm}
\usepackage{algorithmic}
\usepackage{diagbox}
\usepackage{booktabs}
\usepackage{multirow}
\usepackage[most]{tcolorbox}
\usepackage{enumitem}
\usepackage{afterpage}
\usepackage{placeins}

\definecolor{darkblue}{HTML}{1F33B4}
\definecolor{slightblue}{HTML}{77AFDF}
\tcbuselibrary{listings,skins,breakable}
\newtcblisting{templatebox}[1]{
    enhanced,
    colback=white,
    colframe=slightblue,
    colbacktitle=slightblue,
    coltitle=white,
    fonttitle=\bfseries,
    title=#1,
    halign=flush left,
    arc=2.5mm,
    boxrule=1pt,
    drop fuzzy shadow={gray!50!white},
    boxsep=0.8mm,
    left=1.5mm,
    right=1.5mm,
    top=1mm,
    bottom=2mm,
    listing only,
    listing options={
        breaklines=true,
        breakautoindent=false,
        breakindent=0pt,
        aboveskip=0pt,
        belowskip=0pt,
        columns=fullflexible,
        keepspaces=true,
        showstringspaces=false,
        tabsize=2
    }
}

\title{SkillForge: Evolving Verifiable Skills for Reinforcement Learning Agents}

\author{
  Shidong Yang\textsuperscript{*},
  Ziyu Ma\textsuperscript{*},
  Tongwen Huang,
  Xucong Wang, \\
  \textbf{Renda Li},
  \textbf{Yiming Hu},
  \textbf{Yong Wang\textsuperscript{\textdagger}},
  \textbf{Xiangxiang Chu} \\
  AMAP, Alibaba Group
  }

\begin{document}
\maketitle
\begingroup
\renewcommand{\thefootnote}{\fnsymbol{footnote}}
  \footnotetext[1]{Equal contribution.}
  \footnotetext[2]{Project lead and corresponding author.}
\endgroup

\begin{abstract}
Large language model (LLM) agents are trained with reinforcement learning (RL) for complex decision-making tasks. However, most RL-trained agents remain episodic and cannot accumulate reusable knowledge across episodes. Recent skill-based approaches, such as \textsc{SkillRL}, attempt to address this issue by extracting skills from raw trajectories, but treat the skill bank as an append-only repository without verifying whether stored skills remain effective. In this paper, we propose \textsc{SkillForge}, a framework for continuous skill evolution that enables skills to be verified and refined through environment interaction. By making skill usage explicit during agent interaction, RL can directly optimize both environment actions and skill invocation decisions. \textsc{SkillForge} further introduces evidence-based skill verification and multi-pathway skill induction, allowing the skill bank to continuously grow while maintaining its quality. Extensive experiments on ALFWorld, WebShop, and AppWorld show that \textsc{SkillForge} consistently outperforms \textsc{SkillRL}, demonstrating the effectiveness of continuously verified skills in training stronger LLM agents.
\end{abstract}

\section{Introduction}
\label{sec:intro}

The rapid advancement of large language models (LLMs)~\cite{liu2024deepseek,yang2025qwen3} has driven the development of LLM-based agents~\cite{yao2022react,shinn2023reflexion}, which are now widely used in applications such as web navigation~\cite{geminigui,openaigui}, deep research, and personal assistance~\cite{openaideepresearch,geminideepresearch,team2025tongyi}. Reinforcement learning (RL)~\cite{guo2025deepseek,sun2024llm, ji2025tree,chu2025gpg} has become the dominant approach for training such agents, enabling them to acquire adaptive behaviors in open-ended environments. However, most RL-trained agents~\cite{li2025deepagent, mai2025agent, lin2025comprehensive} remain episodic: each episode starts from scratch without retaining reusable knowledge from past successes or failures~\cite{zhang2025memevolve}. As a result, agents often have to rediscover effective behaviors repeatedly, leading to inefficient learning. This raises a fundamental question: \textit{how can agents accumulate and reuse knowledge across episodes?}

\begin{figure}[t]
  \includegraphics[width=\columnwidth]{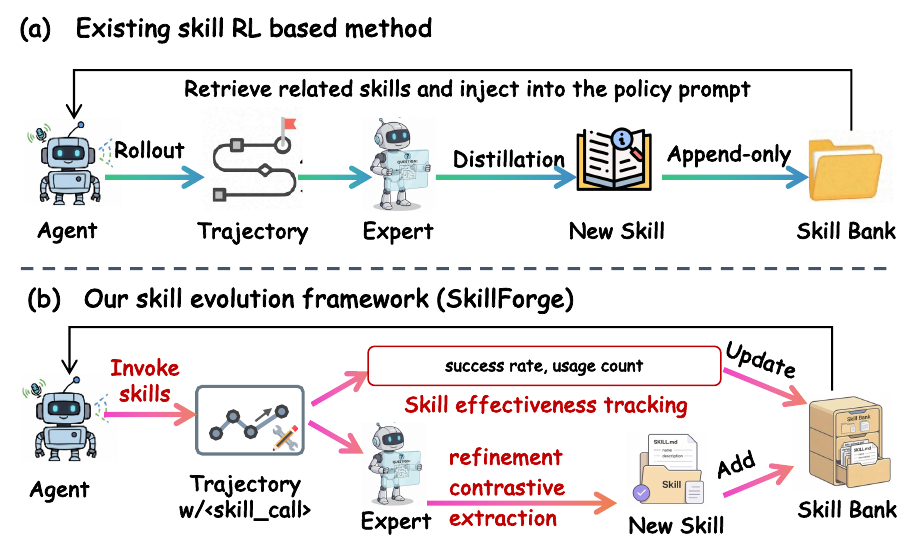}
  \caption{(a)~Existing methods distill trajectories into skills and append them to the bank, but skills are injected in bulk without explicit usage tracking or quality control. (b)~\textsc{SkillForge} introduces explicit skill calling with per-skill effectiveness tracking to update existing skills, and multi-pathway induction to continuously add new skills to the evolving bank.}
  \label{fig:comparative_methods}
\end{figure}

A common solution is to equip agents with external memory that stores past trajectories for future reference~\citep{shinn2023reflexion,zhao2024expel,chhikara2025mem0}. However, raw trajectories are often long, redundant, and noisy~\citep{chhikara2025mem0}, making them difficult to reuse effectively. Even with compression or online updates~\citep{zhang2025memevolve,zhang2026memrl}, existing memory-based methods mainly rely on recalling past episodes rather than extracting reusable knowledge. Instead of memorizing what happened in previous trajectories, a more effective strategy is to distill experience into skills, which are compact decision principles that capture what to do, why it works, and when to apply it~\citep{anthropic2025agentskills}. Such skills are concise, reusable across episodes, and easy to retrieve during decision making. Recent work such as \textsc{SkillRL}~\citep{xia2026skillrl} demonstrates the effectiveness of this direction by extracting skills from raw trajectories and injecting them into the policy prompt (as shown in Fig.~\ref{fig:comparative_methods}(a)), achieving substantial improvements over both vanilla RL and memory-based approaches.

Despite these improvements, \textsc{SkillRL} and related approaches treat the skill bank as an append-only skill repository in which skills are continuously added but rarely examined~\citep{xia2026skillrl}. In practice, skills are written once and then repeatedly injected into the prompt without verifying whether they are actually useful. This design leads to three important limitations. First, skill usage is not observable. Skills are inserted in bulk, and it is unclear whether the agent truly relies on a specific skill during decision making. Second, skill effectiveness is difficult to attribute. Without explicit usage signals, it is hard to determine whether a successful trajectory is caused by a particular skill or simply coincides with it. Third, skill quality is not controlled. Incorrect or outdated skills may remain in the skill bank indefinitely, gradually polluting the knowledge base.

To address these issues, we propose \textsc{SkillForge}, a skill evolution framework that enables skills to be continuously verified through environment interaction (Fig.~\ref{fig:comparative_methods}(b)). The key idea is to make skill usage explicit during agent interaction. Instead of injecting full skill descriptions into the prompt, \textsc{SkillForge} provides a compact catalog and allows the agent to invoke skills on demand using structured tags. Each invocation becomes a discrete event in the trajectory, making skill usage directly observable and allowing RL to reinforce useful skills while suppressing ineffective ones. \textsc{SkillForge} further introduces evidence-based skill verification. For each skill, the framework tracks statistics such as success rate and usage count, which are aggregated into an underperformance score used for skill review. High-scoring skills show stronger evidence of underperformance and are prioritized for LLM-based reflexion, and they are revised when needed to maintain skill quality. Finally, \textsc{SkillForge} introduces a multi-pathway skill induction mechanism that synthesizes new skills from successful trajectories, failed attempts, and contrastive outcome analysis, enabling the skill bank to continuously grow while improving its quality. We evaluate \textsc{SkillForge} on ALFWorld~\cite{shridharalfworld}, WebShop~\cite{yao2022webshop}, and AppWorld~\cite{trivedi2024appworld}. Without cold-start initialization, \textsc{SkillForge} improves over \textsc{SkillRL} by 6.3\% on average, demonstrating that continuously verified skills lead to stronger agents.

Our contributions can be summarized as follows:
\begin{itemize}[leftmargin=*]
\item We propose \textsc{SkillForge}, a continuous skill evolution framework that enables skills to be verified and refined through environment interaction.
\item We design an explicit skill calling strategy that makes skill usage observable and allows RL to jointly optimize actions and skill invocations.
\item Extensive experiments on three diverse benchmarks show that \textsc{SkillForge} consistently outperforms existing skill-based baselines. \end{itemize}

\section{Related Work}
\label{sec:related}

\noindent \textbf{LLM Agents.}
Recent progress in large language models has led to a surge of autonomous agents that can reason, act, and interact with external environments~\citep{wei2026agentic}. Representative frameworks such as ReAct~\cite{yao2022react} and Reflexion~\cite{shinn2023reflexion} augment agent behavior with interleaved reasoning or self-reflection, while AutoGen~\cite{wu2024autogen} and CAMEL~\cite{li2023camel} extend this paradigm to multi-agent collaboration and tool use. Despite their strong empirical performance, most existing agents are still built on in-context learning~\cite{dong2024survey} and remain fundamentally episodic, with little ability to accumulate reusable knowledge across interactions.

\noindent \textbf{Memory in Agents.}
To improve long-horizon decision making, many agent systems introduce external memory~\cite{hu2025memory}. Early approaches mainly rely on retrieval-augmented generation or direct storage of past trajectories~\cite{wangvoyager,chhikara2025mem0,zhang2025g,wang2024agent}. More recent work moves toward compressing experience into summaries, tips, or higher-level reflections~\cite{wang2025mirix,tang2025agent,fang2025memp,zhao2024expel,ouyang2025reasoningbank,wei2025evo}. Although these methods improve memory efficiency, they retain noisy or weakly grounded information, making it difficult to extract reusable knowledge that can benefit future decisions.

\begin{figure*}[t]
  \includegraphics[width=\textwidth]{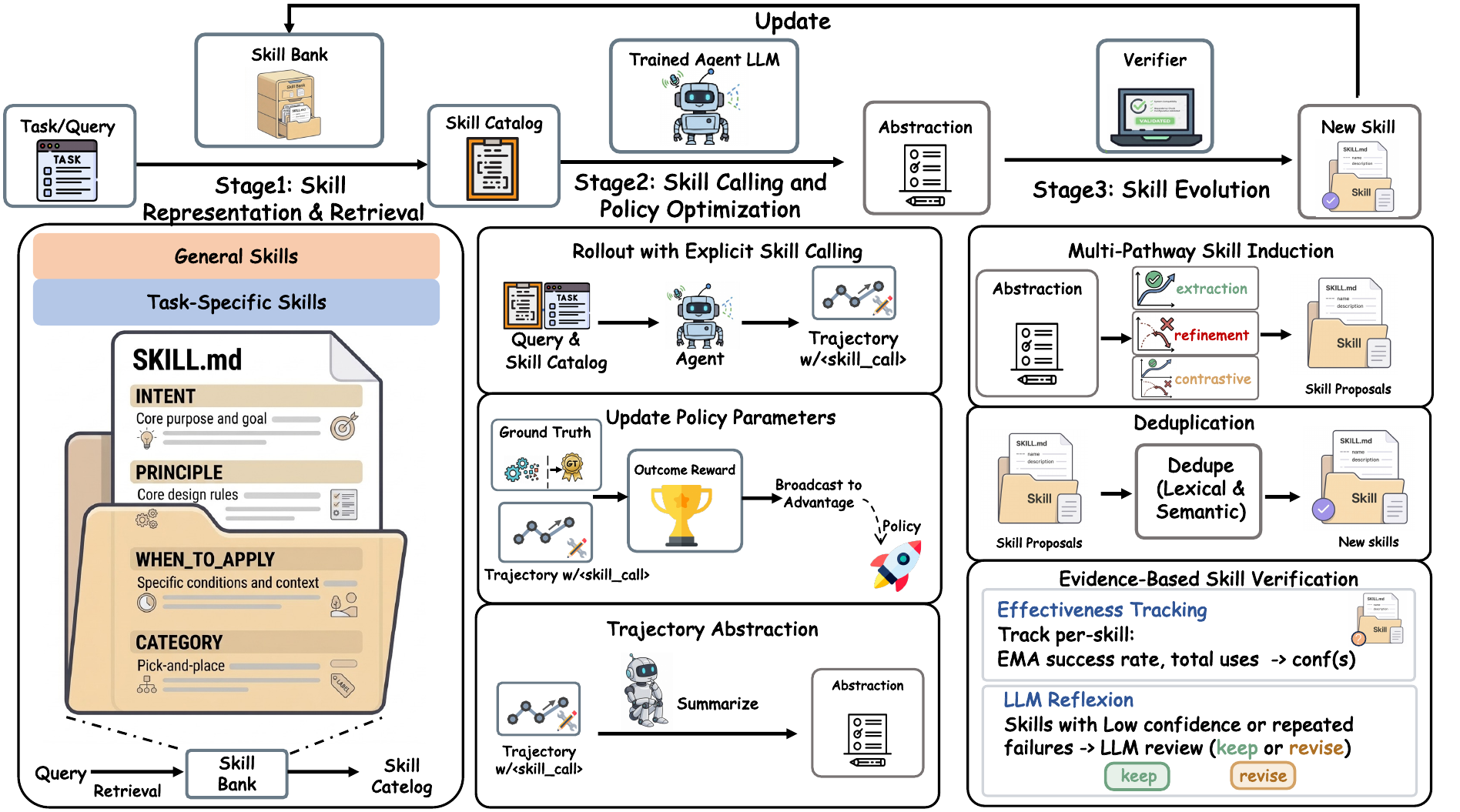}
\caption{
Overview of the \textsc{SkillForge} framework. Skills are retrieved from the skill bank and injected as a compact catalog into the agent's prompt (\S\ref{sec:skill_repr}). During rollout, the agent explicitly invokes skills via structured tags, and the policy is optimized with GRPO (\S\ref{sec:skill_calling}). Rollout trajectories then drive multi-pathway skill induction and evidence-based verification to continuously evolve the skill bank (\S\ref{sec:skill_evolution}).
}
\label{fig:framework}
\end{figure*}

\noindent \textbf{Skill Learning for Agents.}
Beyond generic memory, recent studies increasingly explore reusable skills as a more structured form of experience abstraction~\cite{anthropic2025agentskills,gao2025survey,xia2025agent0,liu2025agent0,xia2026skillrl, ma2026skillclaw}. This direction is related to continual learning~\cite{parisi2019continual} and is also connected to reinforcement learning, which has been widely used for model alignment and reasoning optimization~\cite{schulman2017proximal,ouyang2022training,shao2024deepseekmath}. However, skill learning in open-ended agent environments remains underexplored, especially when skills must be updated according to long-horizon interaction outcomes. This motivates studying skills not as static memory items, but as reusable capabilities that can improve over time.

\section{Method}
\label{sec:method}

We propose \textsc{SkillForge}, a framework that maintains a continuously verifiable skill bank alongside the agent's policy during RL training. In this section, we first describe the skill representation and retrieval process in Section~\ref{sec:skill_repr}. We then present the rollout and policy optimization process with explicit skill calling in Section~\ref{sec:skill_calling}. Finally, Section~\ref{sec:skill_evolution} introduces skill induction, verification, and update based on environment interaction. The overall framework is illustrated in Fig.~\ref{fig:framework}.

\subsection{Skill Representation and Retrieval}
\label{sec:skill_repr}

\paragraph{Problem Setup.}
We consider an LLM agent with policy $\pi_\theta$ interacting with an environment $\mathcal{E}$ over multi-step episodes. Each task is specified by a description $d$. At step $t$, the agent observes $o_t$, produces action $a_t$, and receives the next observation $o_{t+1}$, forming a trajectory $\tau = (o_0, a_0, \ldots, o_T)$ with binary reward $r(\tau) \in \{0, 1\}$. We augment the agent with a \emph{skill bank} $\mathcal{B} = \mathcal{B}_g \cup \bigcup_{k=1}^{K} \mathcal{B}_k$, where $\mathcal{B}_g$ contains general skills and each $\mathcal{B}_k$ corresponds to a task type and contains its associated skills. The joint objective is:
\begin{equation}
\label{eq:objective}
\max_{\theta,\;\mathcal{B}} \;\; \mathbb{E}_{d \sim \mathcal{D}} \Big[\, \mathbb{E}_{\tau \sim \pi_\theta(\cdot \mid d,\, \mathcal{B})} \big[\, r(\tau) \,\big] \,\Big]\,.
\end{equation}

\paragraph{Skill Schema.}
Each skill $s \in \mathcal{B}$ is a structured knowledge unit. It consists of a \emph{title} as the callable identifier, an \emph{intent} describing its purpose, a \emph{principle} encoding the core decision strategy, \emph{applicability conditions} specifying when to apply it, a \emph{category} label (\texttt{general} or task type $k$), and a \emph{status} flag for tracking whether the skill is active or under revision. This design keeps each skill self-contained: the agent can judge its relevance from the intent and access the full content only after explicit invocation.

\paragraph{Skill Bank Initialization.}
We construct the initial skill bank $\mathcal{B}_0$ by rolling out the initial instruction-tuned policy model in the target environment and distilling the collected trajectories through a teacher LLM $M_T$, following \textsc{SkillRL}~\citep{xia2026skillrl}. Successful trajectories are distilled into strategic patterns, while failed ones are synthesized into concise corrective lessons. The resulting skills are organized into a two-level hierarchy: \emph{general skills} $\mathcal{B}_g$ capture universal strategies applicable across all task types (e.g., systematic exploration, precondition checking), while \emph{task-specific skills} $\mathcal{B}_k$ encode specialized knowledge for each task category $k$ (e.g., domain-specific action sequences, common failure modes). This initial bank serves as the starting point for subsequent evolution during RL training.

\paragraph{Retrieval and Catalog Injection.}
At the beginning of each episode, the system retrieves a subset of skills from $\mathcal{B}_g \cup\mathcal{B}_k$ using embedding-based retrieval, which ranks all skills by semantic similarity to the task description $d$ and selects the top-$K$:

\begin{equation}
\label{eq:retrieval}
\mathcal{S}_{\text{ret}} = \underset{s \,\in\, \mathcal{B}}{\text{Top-}K}\; \cos(\mathbf{e}_d,\, \mathbf{e}_s)\,,
\end{equation}
where $\mathbf{e}_d$ and $\mathbf{e}_s$ are the embeddings of the task description and skill intent, respectively. The retrieved skills are formatted into a compact catalog containing only the title and a one-line intent summary, and then appended to the system prompt. This design keeps the prompt compact regardless of the bank size, while the full skill content, including the principle and applicability conditions, is revealed only after explicit skill calling (\S\ref{sec:skill_calling}).

\subsection{Skill Calling and Policy Optimization}
\label{sec:skill_calling}

\paragraph{Rollout with Explicit Skill Calling.}
For each task $d$, the retrieved skill catalog $\mathcal{S}_{\text{ret}}$ is appended to the prompt together with the task query. During interaction with the environment, the agent's output at each step consists of an environment action $a_t^{\text{env}}$ and an optional skill invocation $c_t$:
\begin{equation}
\label{eq:skill_call}
a_t = \bigl(a_t^{\text{env}},\; c_t\bigr), \quad c_t \in \{\varnothing\} \cup \mathcal{S}_{\text{ret}}\,,
\end{equation}
where $c_t$ selects a skill from the retrieved catalog or $\varnothing$ if no skill is invoked. In practice, the agent emits a structured \texttt{<skill\_call>}\textsc{name}\texttt{</skill\_call>} tag within its response. The framework resolves the called name against the catalog and returns the full structured skill content (intent, principle, and applicability conditions) as feedback in the next observation:
\begin{equation}
\label{eq:skill_obs}
\scalebox{0.90}{$
o_{t+1} = \begin{cases}
\mathcal{E}(a_t^{\text{env}}) \;\oplus\; \text{Content}(c_t)\,, & \text{if } c_t \neq \varnothing\,, \\
\mathcal{E}(a_t^{\text{env}})\,, & \text{otherwise}\,,
\end{cases}
$}
\end{equation}
where $\oplus$ denotes concatenation. This design treats skill invocation as a discrete, traceable action within the trajectory, making skill usage both observable and attributable. Since the calling tag is part of the generated token sequence, RL directly optimizes when and which skills the agent invokes. Per-skill usage events further enable effectiveness tracking in Section~\ref{sec:skill_evolution}.
\paragraph{Policy Update.}
The agent samples $G$ trajectories $\{\tau^{(i)}\}_{i=1}^{G} \sim \pi_\theta(\cdot \mid d, \mathcal{S}_{\text{ret}})$ with skill calling events recorded. Each trajectory is scored against the ground-truth outcome reward $R_i = r(\tau^{(i)})$, from which we compute the group-relative advantage $\hat{A}_i = (R_i - \bar{R}) / \sigma_R$. The policy is optimized via GRPO~\citep{guo2025deepseek}:
\begin{equation}
\label{eq:grpo}
\scalebox{0.84}{$
\begin{aligned}
J(\theta) = \mathbb{E}\Bigg[
\frac{1}{G}\sum_{i=1}^{G}
\min\!\Big(\rho_i \hat{A}_i,\;
\text{clip}(\rho_i,1{-}\epsilon,1{+}\epsilon)\hat{A}_i\Big)
\\
- \beta\, D_{\text{KL}}\!\bigl(\pi_\theta \| \pi_{\text{ref}}\bigr)
\Bigg]
\end{aligned}
$}
\end{equation}
where $\rho_i = \pi_\theta / \pi_{\theta_{\text{old}}}$ is the importance ratio, $\epsilon$ is the clipping range, and $\pi_{\text{ref}}$ is a fixed reference policy. Since the \texttt{<skill\_call>} tag is part of the generated token sequence, the same objective jointly optimizes actions and skill invocation decisions.

\paragraph{Trajectory Abstraction.}
Each rollout trajectory is summarized by an LLM into a concise abstraction preserving key decisions, skill calls, and outcomes while discarding redundant observations. Trajectories are partitioned into successful ($\mathcal{T}^+$) and failed ($\mathcal{T}^-$) sets according to reward and stored in per-task-type buffers, which serve as structured input for skill induction.

\begin{table*}[!t]
\centering

\resizebox{\textwidth}{!}{
\begin{tabular}{lccccccc|cc|cc}
\toprule
\multirow{2}{*}{\textbf{Method}} & \multicolumn{7}{c|}{\textbf{ALFWorld}} & \multicolumn{2}{c|}{\textbf{WebShop}} & \multicolumn{2}{c}{\textbf{AppWorld}} \\
 & Pick & Look & Clean & Heat & Cool & Pick2 & All & Score & Succ. & TGC & SGC \\
\midrule
\rowcolor{gray!15} \multicolumn{12}{l}{\textit{Closed-source LLMs}} \\
GPT-4o & 75.3 & 60.8 & 31.2 & 56.7 & 21.6 & 49.8 & 48.0 & 31.8 & 23.7 & \underline{48.8} & \underline{32.1} \\
Gemini-2.5-Pro & \underline{92.8} & 63.3 & 62.1 & 69.0 & 26.6 & 58.7 & 60.3 & 42.5 & 35.9 & -- & -- \\
\midrule
\textit{Qwen2.5-7B-Instruct} & & & & & & & & & & & \\
Qwen2.5 & 33.4 & 21.6 & 19.3 & 6.90 & 2.80 & 3.20 & 14.8 & 26.4 & 7.80 & 0.59 & 0.00 \\
\rowcolor{gray!15} \multicolumn{12}{l}{\textit{Prompt-based Agentic or Memory-based Methods}} \\
ReAct$^*$ & 48.5 & 35.4 & 34.3 & 13.2 & 18.2 & 17.6 & 31.2 & 46.2 & 19.5 & 0.59 & 0.00 \\
Reflexion$^*$ & 62.0 & 41.6 & 44.9 & 30.9 & 36.3 & 23.8 & 42.7 & 58.1 & 28.8 & 1.19 & 0.00 \\
Mem0 & 54.0 & 55.0 & 26.9 & 36.4 & 20.8 & 7.69 & 33.6 & 23.9 & 2.00 & -- & -- \\
SimpleMem & 64.5 & 33.3 & 20.0 & 12.5 & 33.3 & 3.84 & 29.7 & 33.2 & 8.59 & -- & -- \\
\midrule
\rowcolor{gray!15} \multicolumn{12}{l}{\textit{RL-based Methods}} \\
RLOO$^*$ & 87.6 & 78.2 & 87.3 & 81.3 & 71.9 & 48.9 & 75.5 & 80.3 & 65.7 & -- & -- \\
GRPO$^*$ & 90.8 & 66.1 & 89.3 & 74.7 & 72.5 & 64.7 & 77.6 & 79.3 & 66.1 & 17.9 & 3.57 \\
\midrule
\rowcolor{gray!15} \multicolumn{12}{l}{\textit{Memory-Augmented RL-based Methods}} \\
Mem0+GRPO & 78.1 & 54.8 & 56.1 & 31.0 & 65.0 & 26.9 & 54.7 & 58.1 & 37.5 & -- & -- \\
SimpleMem+GRPO & 89.5 & 36.3 & 60.0 & 50.0 & 64.9 & 26.3 & 62.5 & 67.8 & 46.9 & -- & -- \\

SkillRL & \underline{97.9} & 71.4 & \underline{90.0} & 90.0 & \textbf{95.5} & 87.5 & 89.9 & 85.2 & 72.7 & 19.0 & 5.36 \\
\midrule
\textsc{Ours}(Qwen2.5-7B) & \textbf{100} & \textbf{100} & \textbf{92.6} & \textbf{100} & 80.0 & \underline{91.7} & \underline{93.6} & \underline{89.8} & 83.0 & 23.8 & 14.3 \\
\textsc{Ours}(Qwen3-4B) & 94.3 & \underline{92.3} & \textbf{92.6} & 81.3 & 84.0 & 79.2 & 87.9 & \textbf{90.8} & \textbf{84.0} & 44.6& 30.4 \\
\textsc{Ours}(Qwen3-30B-A3B) & \textbf{100} & 84.6 & \textbf{92.6} & \underline{93.8} & \underline{92.0} & \textbf{95.8} & \textbf{94.3} & \textbf{90.8} & \underline{83.8} & \textbf{59.5} & \textbf{37.5} \\
\bottomrule
\end{tabular}
}
\caption{Performance on ALFWorld, WebShop, and AppWorld. For ALFWorld, we report the average success rate (\%) for each subtask as well as the overall result. For WebShop, we report both the average score and the average success rate (\%). For AppWorld, we report Task Goal Completion (TGC) and Scenario Goal Completion (SGC). $^*$ denotes the results replicated from~\citep{feng2025group}.}
\label{tab:performance}
\end{table*}

\subsection{Skill Evolution}
\label{sec:skill_evolution}

The skill bank evolves continuously through two processes. \emph{Induction} adds new skills from rollout trajectories, while \emph{verification} evaluates and refines existing skills using usage evidence.

\paragraph{Multi-Pathway Skill Induction.}
Every $I$ training steps, a teacher LLM $M_T$ synthesizes new skills from the trajectory abstractions collected in Section~\ref{sec:skill_calling}. Depending on the trajectory buffer, one of three induction pathways is selected. \emph{Extraction} extracts generalizable strategies from successful trajectories $\mathcal{T}^+$. \emph{Refinement} identifies recurring failure patterns from $\mathcal{T}^-$ and formulates corrective strategies. \emph{Contrastive} analysis pairs successes with failures to identify decisive behavioral differences. The teacher synthesizes new skills as:
\begin{equation}
\label{eq:skill_synthesis}
\mathcal{S}_{\text{new}} = M_T\!\bigl(\mathcal{T}^+, \mathcal{T}^-, \mathcal{B},\; m\bigr),
\end{equation}
where the mode $m$ is \emph{contrastive} when both $|\mathcal{T}^+|,|\mathcal{T}^-|>0$, \emph{extraction} when only successes are available, and \emph{refinement} when only failures exist. The existing bank $\mathcal{B}$ is provided as context to avoid generating redundant skills. New skills are deduplicated via lexical matching and semantic similarity before being added to the bank.

\paragraph{Evidence-Based Skill Verification.}
Explicit skill calling (Section~\ref{sec:skill_calling}) makes each skill invocation a traceable event tied to an episode outcome. This enables fine-grained effectiveness tracking. We maintain per-skill statistics updated after every training step. Specifically, each time skill $s$ is invoked in an episode with outcome $r \in \{0,1\}$, its exponential moving average (EMA) success rate is updated as:
\begin{equation}
\label{eq:ema}
\hat{p}_s \leftarrow \alpha \cdot r + (1 - \alpha) \cdot \hat{p}_s\,,
\end{equation}
where $\alpha$ is the smoothing factor. We additionally track total uses $n_s$. These statistics are combined into an underperformance score:
\begin{equation}
\label{eq:confidence}
\mathrm{conf}(s) = (1 - \hat{p}_s) \cdot \bigl(1 - 0.5^{\,n_s / h}\bigr)\,,
\end{equation}
where $h$ is the usage half-life. A skill with low success rate and high usage receives a high underperformance score, indicating stronger evidence that it should be prioritized for \emph{reflexion}. An LLM reviews the skill definition against its recent usage contexts and returns a verdict of \textbf{keep} or \textbf{revise} (rewriting the principle and applicability conditions). This process ensures that skills that no longer contribute are revised in a timely manner, preventing knowledge decay and maintaining the overall reliability of the skill bank.

\paragraph{Training Procedure.}
Algorithm~\ref{alg:skillforge} summarizes the complete loop. Each training step consists of skill-augmented rollout (Section~\ref{sec:skill_calling}) followed by policy update. Every $I$ steps, the skill bank is updated: new skills are induced and deduplicated, existing skills are verified through effectiveness tracking and reflexion, and the updated bank is immediately available to subsequent rollouts. This ensures that skills are continuously generated, tested, and refined through environment interactions.

\begin{figure*}[t]
\centering
\includegraphics[width=2\columnwidth]{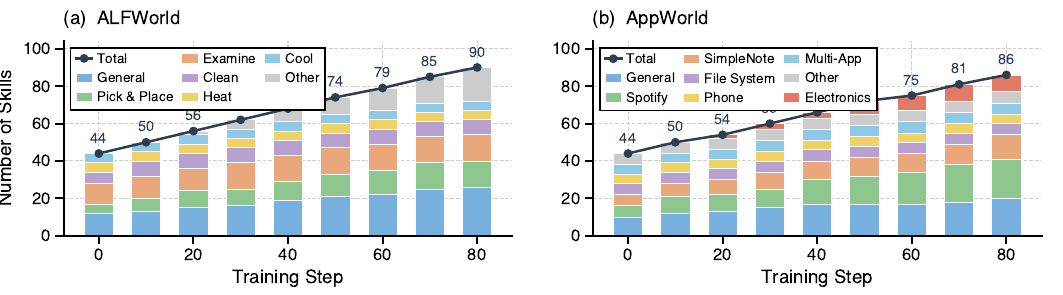}
\caption{Evolution of the skill bank during training on ALFWorld and AppWorld. 
The y-axis shows the number of active skills, grouped into general and task-specific categories. New skills are continuously induced via trajectory analysis, while potentially underperforming ones are reviewed and revised when needed through reflexion, resulting in steady but controlled growth.}
\label{fig:skill_evolution}
\end{figure*}

\begin{table}[t]
\centering

\small
\begin{tabular}{lcc}
\toprule
\textbf{Method} & \textbf{ALFWorld} & \textbf{AppWorld} \\
\midrule
\textsc{SkillForge} & \textbf{87.9} & \textbf{44.6} \\
\midrule
\rowcolor{gray!10} \multicolumn{3}{l}{\textit{Skill Calling}} \\
\quad w/o Explicit Calling & 77.9 & 33.3 \\
\quad w/o Skill Bank & 79.3 & 34.5 \\
\midrule
\rowcolor{gray!10} \multicolumn{3}{l}{\textit{Skill Induction}} \\
\quad w/o Multi-Pathway & 82.1  & 36.9 \\
\quad w/o Deduplication & 86.4 & 38.7 \\
\midrule
\rowcolor{gray!10} \multicolumn{3}{l}{\textit{Skill Verification}} \\
\quad w/o Effectiveness Tracking & 83.6 & 36.3 \\
\quad w/o LLM Reflexion & 82.1 & 39.3 \\

\bottomrule
\end{tabular}
\caption{Ablation study on Qwen3-4B. We report ALFWorld success rate (\%) and AppWorld TGC (\%).}
\label{tab:ablation}
\end{table}

\section{Experiments}
\label{sec:experiment}

\subsection{Experimental Setup}

We evaluate \textsc{SkillForge} on three benchmarks: ALFWorld~\cite{shridharalfworld}, WebShop~\cite{yao2022webshop}, and AppWorld~\cite{trivedi2024appworld}.
For ALFWorld, we report per-subtask and overall success rate (\%). For WebShop, we report average score and success rate (\%). For AppWorld, we report Task Goal Completion (TGC) and Scenario Goal Completion (SGC).

\subsection{Implementation Details}
We implement all experiments with the VeRL framework~\cite{shao2024verl}. Specifically, Qwen2.5-7B-Instruct and Qwen3-4B-Instruct are trained on one node with 8 NVIDIA H20 GPUs, while Qwen3-30B-A3B-Instruct is trained on 16 H20 GPUs.
We use GRPO with a constant learning rate of $1\mathrm{e}{-6}$, $G{=}8$ samples per prompt, and KL coefficient $1\mathrm{e}{-3}$. Rollout temperature is 0.9. The skill bank is updated every $I{=}5$ training steps using Qwen3-Max as the teacher LLM $M_T$.
For skill retrieval, we use Qwen3-Embedding-0.6B to encode task descriptions and serialized skill representations.

\begin{figure*}[t]
\centering
\includegraphics[width=2\columnwidth]{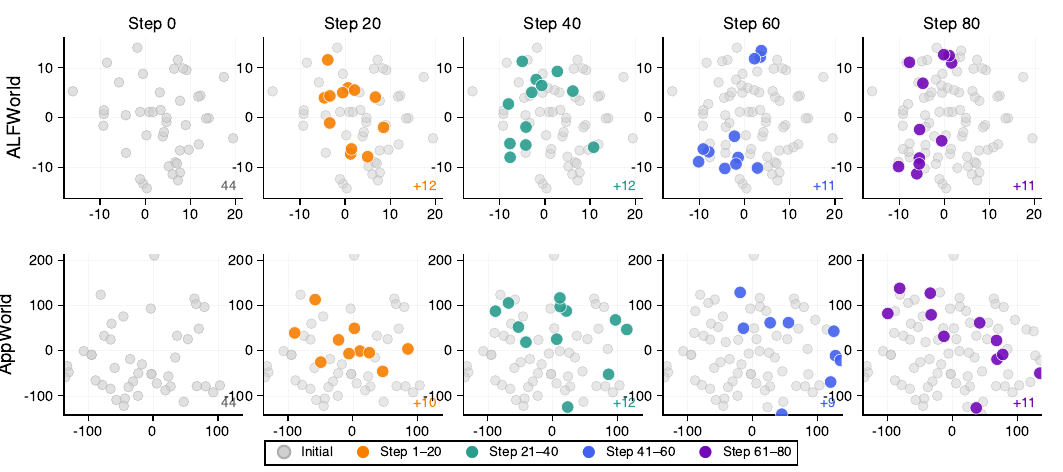}
\caption{t-SNE visualization of skill intent embeddings at different training steps on ALFWorld (top) and AppWorld (bottom). Gray points denote the initial skill bank, while colored points represent skills induced during training. As training progresses (Step 20, 40, 60, and 80), newly generated skills gradually expand into previously unexplored regions of the embedding space, indicating the progressive diversification of the skill bank through trajectory-driven induction and reflexion-based refinement.}

\label{fig:skill_distribution}
\end{figure*}

\subsection{Main Results}

\paragraph{Overall Performance.}
Table~\ref{tab:performance} reports the main results. \textsc{SkillForge} consistently outperforms all baselines.
With Qwen2.5-7B, it achieves 93.6 on ALFWorld and 89.8/83.0 (score/success) on WebShop, improving over GRPO by +16.0 and +10.5 score (+16.9 success), respectively.
On the more challenging AppWorld, it further increases TGC/SGC from 17.9/3.57 to 23.8/14.3, demonstrating strong improvements on multi-step tasks.

\paragraph{Comparison with Skill-based Methods.}
Compared with the strongest baseline \textsc{SkillRL}, \textsc{SkillForge} improves performance under the same backbone by +3.7 on ALFWorld (93.6 vs.\ 89.9) and +10.3 success on WebShop (83.0 vs.\ 72.7).
The gap becomes larger on AppWorld, where TGC increases from 19.0 to 23.8 and SGC nearly triples (5.36 $\rightarrow$ 14.3), highlighting the benefit of continuous skill refinement.

\paragraph{Scaling Across Models.}
\textsc{SkillForge} benefits all model scales.
Notably, Qwen3-4B already achieves strong performance (87.9 ALFWorld, 84.0 WebShop success), approaching or surpassing Qwen2.5-7B with \textsc{SkillRL}.
Scaling to Qwen3-30B-A3B further yields the best results (94.3 ALFWorld, 59.5 AppWorld TGC).

\subsection{Analysis}

\noindent\textbf{Effect of Skill Evolution Components.}
Table~\ref{tab:ablation} studies the contribution of each component on Qwen3-4B.
Removing explicit skill calling or the skill bank leads to large drops (87.9$\rightarrow$77.9 and 79.3 on ALFWorld), confirming the importance of external skill retrieval.
Within skill induction, multi-pathway induction is the most critical component, while deduplication is important on AppWorld, where TGC drops from 44.6 to 38.7.
For skill verification, both effectiveness tracking and LLM reflexion improve performance, with the former contributing more substantially.
Overall, skill calling, induction, and verification all provide complementary gains.

\noindent\textbf{Skill Bank Evolution.}
Fig.~\ref{fig:skill_evolution} illustrates how the skill bank evolves during training on ALFWorld and AppWorld.
The total number of active skills grows steadily (44$\rightarrow$90 on ALFWorld and 44$\rightarrow$86 on AppWorld) as the agent encounters diverse task failures.
General skills increase gradually, while task-specific skills expand to address environment-specific requirements.
This controlled growth suggests that \textsc{SkillForge} continuously induces useful skills while revising unreliable ones through reflexion, maintaining a compact yet expressive skill bank.
\begin{table}[t]
\centering

\small
\begin{tabular}{@{}lcc@{}}
\toprule
\textbf{Skill Bank Source} & \textbf{ALFWorld} & \textbf{AppWorld} \\
\midrule
No skill & 26.4 & 26.8 \\
\midrule
Qwen3-4B @ init (step 0) & 27.9 & 27.4 \\
Qwen3-4B @ step 20 & 28.6 & 28.6 \\
Qwen3-4B @ step 40 & 28.6 & \textbf{31.5} \\
Qwen3-4B @ step 80 & 32.9 & \textbf{31.5} \\
\midrule
Qwen3-30B-A3B self-evolved & \textbf{34.3} & 30.4 \\
\bottomrule
\end{tabular}
\caption{Skill transferability across model scales. Skill banks produced by Qwen3-4B-Instruct at different training stages are directly applied to Qwen3-30B-A3B-Instruct without further training.}
\label{tab:skill_transfer}
\end{table}

\begin{figure*}[t]
\centering
\includegraphics[width=2\columnwidth]{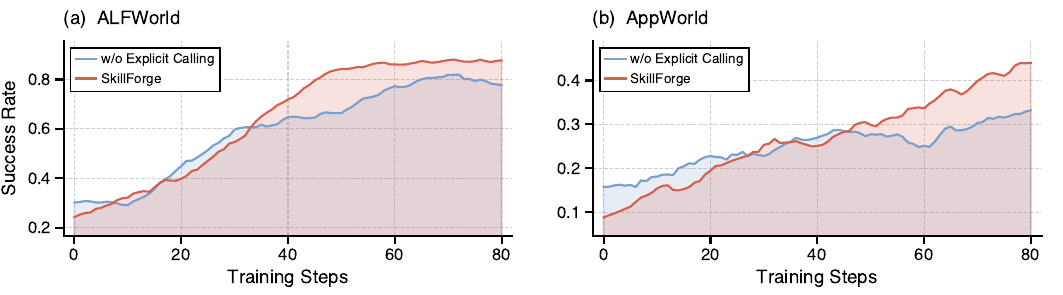}
\caption{Training curves on ALFWorld and AppWorld. \textsc{SkillForge} achieves faster convergence and higher asymptotic performance than the variant without explicit calling.}
\label{fig:training_curve}
\end{figure*}

\begin{figure*}[t]
\centering
\includegraphics[width=\textwidth]{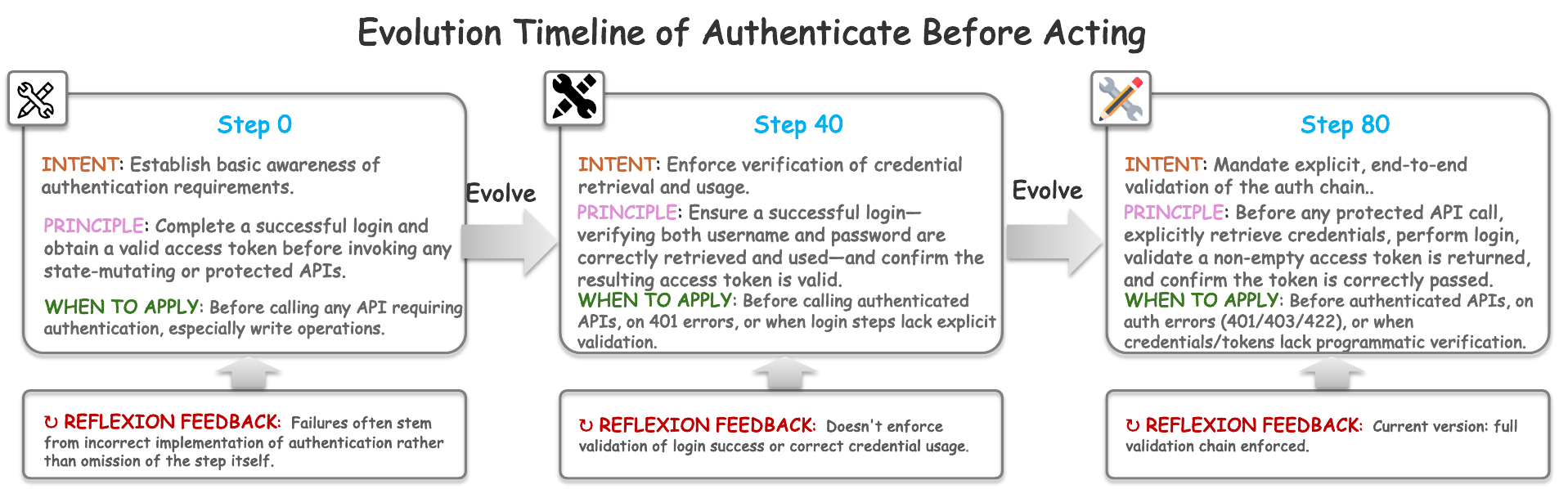}
\caption{Evolution timeline of a representative skill (\textit{Authenticate Before Acting}) across training stages on AppWorld. The figure shows how the skill evolves from Step~0 to Step~40 and Step~80 through trajectory-driven induction and reflexion. The initial version captures a basic authentication strategy. As training progresses, the skill is refined to include explicit credential verification and stronger applicability conditions. Green text marks newly added content, while red text indicates revised or removed parts of the skill definition.}
\label{fig:skill_case}
\end{figure*}

\noindent\textbf{Skill Distribution.}
Fig.~\ref{fig:skill_distribution} visualizes the t-SNE projection of skill intent embeddings at different training steps on both ALFWorld and AppWorld. The initial skill bank (gray points) occupies a limited region of the embedding space. As training progresses, newly induced skills (colored points) gradually expand into new regions, indicating the increasing diversity of the learned strategies. This expansion pattern suggests that trajectory-driven induction continuously introduces skills addressing new task situations, while reflexion-based verification helps maintain the coherence of existing skills. As a result, the skill bank evolves from a small initial set into a richer and more specialized collection of decision strategies.

\noindent\textbf{Skill Transferability.}
Table~\ref{tab:skill_transfer} examines whether evolved skills transfer across model scales by applying Qwen3-4B skill banks to Qwen3-30B-A3B without further training. Later-stage skills consistently outperform earlier ones (32.9 vs.\ 27.9 on ALFWorld, 31.5 vs.\ 27.4 on AppWorld), and even initial skills already improve over the no-skill baseline, confirming that \textsc{SkillForge}'s skills store transferable knowledge rather than model-specific artifacts. Notably, Qwen3-4B step-80 skills match or exceed Qwen3-30B-A3B's self-evolved bank on AppWorld (31.5 vs.\ 30.4), suggesting that a smaller model with sufficient evolution can produce skills competitive with those from a larger model.

\noindent\textbf{Training Dynamics.}
Fig.~\ref{fig:training_curve} shows training curves on ALFWorld and AppWorld. \textsc{SkillForge} initially lags behind the variant without explicit calling due to the overhead of learning skill usage, but quickly surpasses it and achieves both faster convergence and higher final performance. The gap widens in later stages, indicating that continuous skill induction and verification provide compounding benefits during training.

\noindent\textbf{Case Study.}
Fig.~\ref{fig:skill_case} illustrates how a representative skill evolves during training on AppWorld. The initial version (Step~0) captures a coarse strategy that simply requires obtaining a valid authentication token before calling protected APIs. As training progresses, trajectory feedback reveals common failure patterns, such as missing credential validation or incorrect token usage. The reflexion process then revises the skill to incorporate stronger verification rules and clearer applicability conditions. By Step~80, the skill becomes a more precise and structured decision guideline that explicitly enforces end-to-end authentication checks. This example demonstrates how \textsc{SkillForge} continuously refines skills through environment feedback, transforming simple heuristics into more reliable strategies over the course of training.

\section{Conclusion}

We present \textsc{SkillForge}, a framework for continuous skill evolution in reinforcement learning agents. By making skill usage explicit during interaction, \textsc{SkillForge} enables reinforcement learning to optimize both environment actions and skill invocation decisions, while introducing evidence-based skill verification and multi-pathway skill induction to maintain skill quality as the bank grows. Experiments on ALFWorld, WebShop, and AppWorld show that \textsc{SkillForge} consistently outperforms existing approaches such as \textsc{SkillRL} without requiring an SFT initialization stage. These results highlight the importance of continuously verifying and refining skills during training and point toward more adaptive, skill-driven reinforcement learning agents in open-ended environments.
\section*{Limitations}

While \textsc{SkillForge} demonstrates consistent improvements across multiple agent benchmarks, it has several limitations. First, the skill induction and verification processes rely on an external teacher LLM to synthesize and revise skills from trajectory abstractions. The quality of the resulting skill bank therefore depends on the capability of the teacher model. Second, although the verification mechanism helps maintain skill quality, the skill bank may still grow over time as new skills are continuously induced, which could introduce additional retrieval overhead in long training runs. Finally, the explicit skill calling design introduces additional tokens during interaction, which may increase prompt length and inference cost in large-scale deployments.

% \section*{Acknowledgments}

% This document has been adapted
% by Steven Bethard, Ryan Cotterell and Rui Yan
% from the instructions for earlier ACL and NAACL proceedings, including those for
% ACL 2019 by Douwe Kiela and Ivan Vuli\'{c},
% NAACL 2019 by Stephanie Lukin and Alla Roskovskaya,
% ACL 2018 by Shay Cohen, Kevin Gimpel, and Wei Lu,
% NAACL 2018 by Margaret Mitchell and Stephanie Lukin,
% Bib\TeX{} suggestions for (NA)ACL 2017/2018 from Jason Eisner,
% ACL 2017 by Dan Gildea and Min-Yen Kan,
% NAACL 2017 by Margaret Mitchell,
% ACL 2012 by Maggie Li and Michael White,
% ACL 2010 by Jing-Shin Chang and Philipp Koehn,
% ACL 2008 by Johanna D. Moore, Simone Teufel, James Allan, and Sadaoki Furui,
% ACL 2005 by Hwee Tou Ng and Kemal Oflazer,
% ACL 2002 by Eugene Charniak and Dekang Lin,
% and earlier ACL and EACL formats written by several people, including
% John Chen, Henry S. Thompson and Donald Walker.
% Additional elements were taken from the formatting instructions of the \emph{International Joint Conference on Artificial Intelligence} and the \emph{Conference on Computer Vision and Pattern Recognition}.

\bibliography{custom}

% \clearpage
\appendix

\section{Appendix}
\label{sec:appendix}

\subsection{Dataset}
\label{sec:appendix:datasets}

\paragraph{AppWorld.}
AppWorld~\cite{trivedi2024appworld} is a simulated environment for real-world digital service interactions, covering applications such as calendaring, email, music streaming, and social platforms. Agents execute tasks by invoking Python APIs (e.g., ``find the most-liked song in my Spotify playlists''), which typically require multi-step reasoning and cross-application information integration. We report Task Goal Completion (TGC) and Scenario Goal Completion (SGC). TGC is the percentage of individual tasks for which the agent passes all evaluation tests. Each task scenario defines a common underlying task pattern and is instantiated into three independent task variants with different requirements and initial states. SGC is the percentage of scenarios for which the agent successfully completes all three variants. Thus, TGC measures per-task capability, whereas SGC measures whether the agent performs consistently across variations of the same scenario.

\paragraph{ALFWorld.}
ALFWorld~\cite{shridharalfworld} is a text-based interactive environment grounded in household embodied tasks from ALFRED.
An agent must complete long-horizon goals in partially observable rooms by issuing textual actions for navigation, container manipulation, and object interaction, covering tasks such as pick-and-place, examination, cleaning, heating, cooling, and multi-object placement.
Evaluation is based on task success rate: an episode is counted as successful only if the full goal is completed, and we report both per-subtask success rates and the overall average success rate.

\paragraph{WebShop.}
WebShop~\cite{yao2022webshop} is an interactive environment that simulates an e-commerce shopping scenario.
An agent interacts with the environment via two actions, \texttt{search[query]} and \texttt{click[element]}, to complete natural-language shopping requests through product search, attribute filtering, and purchase decision-making.
Evaluation is based on the attribute-matching score between the final selected product and the user's request.

\subsection{Baselines}

We compare \textsc{SkillForge} with four categories of methods.
First, we include closed-source LLMs, including GPT-4o~\cite{openai2024gpt4o} and Gemini-2.5-Pro~\cite{comanici2025gemini}, which represent strong general-purpose reasoning capabilities.
Second, we consider prompt-based and memory-based agents, including ReAct~\cite{yao2022react}, Reflexion~\cite{shinn2023reflexion}, Mem0~\cite{chhikara2025mem0}, and SimpleMem~\cite{liu2026simplemem}, which rely on in-context prompting or external memory without parameter updates.
Third, we evaluate RL-based methods, including RLOO~\cite{ahmadian2024back} and GRPO~\cite{shao2024deepseekmath}, which optimize policies via group-based advantage estimation.
Fourth, we include memory-augmented RL methods, such as Mem0+GRPO and SimpleMem+GRPO~\cite{liu2026simplemem}, which integrate persistent memory into RL training. Finally, we compare against \textsc{SkillRL}~\cite{xia2026skillrl}, the most closely related approach that learns skills from raw trajectories but treats the skill bank as an append-only repository. 
\subsection{Implementation Details}
\label{sec:appendix:setup}

\begin{table}[h]
\centering
\setlength{\tabcolsep}{6pt}
\begin{tabular}{@{}ll@{}}
\toprule
\textbf{Parameter} & \textbf{Value} \\
\midrule
Learning rate & $1\mathrm{e}{-6}$ \\
Group size ($n$) & 8 \\
Training batch size & 32 \\
Optimizer & AdamW \\
Clip ratio low & 0.20 \\
Clip ratio high & 0.28 \\
KL coefficient & $1\mathrm{e}{-3}$ \\
Rollout temperature & 0.9 \\
Evaluation temperature & 0 \\
Max response length & 4096 \\
Reward signal & success $=1$, failure $=0$ \\
\bottomrule
\end{tabular}
\caption{Hyperparameters for RL training.}
\label{tab:appendix:rl-hparams}
\end{table}

We train our agents with GRPO using the VeRL framework~\cite{shao2024verl}. The detailed hyperparameters are summarized in Table~\ref{tab:appendix:rl-hparams}.
We run Qwen2.5-7B-Instruct and Qwen3-4B-Instruct on a single node with $8\times$ NVIDIA H20 GPUs (tensor parallelism${}=1$), and train Qwen3-30B-A3B-Instruct on two nodes, each with $8\times$ H20 GPUs (tensor parallelism${}=2$). Each rollout is truncated to at most 15 environment steps for WebShop and ALFWorld, and 30 steps for AppWorld; trajectories that exceed the step limit are counted as failures. We use Qwen3-Max~\cite{qwen3max} as the expert model.

Across all experiments, we employ embedding-based retrieval, selecting the top 6 general skills and the top 6 task-specific skills for each episode. For evidence-based skill verification, we use an EMA smoothing factor of $\alpha=0.1$ and set the usage half-life to $h=20$.

\subsection{Training Procedure for \textsc{SkillForge}}
\label{sec:appendix:skillforge}

Algorithm~\ref{alg:skillforge} summarizes the training loop of \textsc{SkillForge}. Training alternates between (i) skill-augmented rollouts and (ii) policy optimization. At each step, the agent first retrieves a small, task-relevant subset of skills from the current bank and conditions the rollout on these retrieved skills (Section~\ref{sec:skill_calling}). The collected trajectories are then used to update the policy parameters via Eq.~\eqref{eq:grpo}. Meanwhile, \textsc{SkillForge} maintains online effectiveness statistics for each skill by aggregating skill-calling events (Eq.~\eqref{eq:ema}), enabling the system to estimate whether a skill remains useful under the current policy and environment dynamics.

Every $I$ steps, \textsc{SkillForge} performs a skill-bank maintenance cycle. We first abstract trajectories and split them into successful and unsuccessful sets, $\mathcal{T}^+$ and $\mathcal{T}^-$, based on reward. An expert LLM $M_T$ then induces candidate skills from both $\mathcal{T}^+$ and $\mathcal{T}^-$ (Eq.~\eqref{eq:skill_synthesis}), allowing the bank to grow not only by distilling what worked but also by capturing fixes for common failure modes. Newly induced skills are deduplicated before insertion to control redundancy. For existing skills, we compute a confidence score $\mathrm{conf}(s)$ (Eq.~\eqref{eq:confidence}) from the tracked evidence and flag those whose effectiveness degrades. Flagged skills are subsequently sent to a reflexion stage, where the expert model decides to \textbf{keep} the skill (if evidence is insufficient or noisy) or \textbf{revise} it (if the skill has become stale or misleading). The updated bank $\mathcal{B}$ is immediately used by subsequent rollouts, ensuring skills are continuously generated, tested, and refined through environment interaction.

\begin{algorithm}[t]
\caption{\textsc{SkillForge} Training}
\label{alg:skillforge}
\begin{algorithmic}[1]
\REQUIRE Policy $\pi_\theta$, initial skill bank $\mathcal{B}_0$, expert LLM $M_T$, update interval $I$
\STATE $\mathcal{B} \leftarrow \mathcal{B}_0$; initialize per-task-type trajectory buffers
\FOR{step $= 1$ \TO $N$}
    \FOR{each task $d$ in batch}
        \STATE $\mathcal{S}_{\text{ret}} \leftarrow \text{RetrieveCatalog}(d, \mathcal{B})$ \hfill $\triangleright$ Eq.~\eqref{eq:retrieval}
        \STATE Sample $\{\tau^{(i)}\}_{i=1}^{G} \sim \pi_\theta(\cdot \mid d, \mathcal{S}_{\text{ret}})$ \hfill $\triangleright$ Eq.~\eqref{eq:skill_call}
    \ENDFOR
    \STATE Update $\theta$ via Eq.~\eqref{eq:grpo}
    \STATE LLM-abstract trajectories; partition into $\mathcal{T}^+, \mathcal{T}^-$ by reward
    \STATE Update per-skill statistics from calling events \hfill $\triangleright$ Eq.~\eqref{eq:ema}
    \IF{step $\bmod\; I = 0$}
        \STATE $\mathcal{S}_{\text{new}} \leftarrow M_T(\mathcal{T}^+, \mathcal{T}^-, \mathcal{B}, m)$ \hfill $\triangleright$ Eq.~\eqref{eq:skill_synthesis}
        \STATE Deduplicate $\mathcal{S}_{\text{new}}$; add to $\mathcal{B}$
        \STATE Compute $\mathrm{conf}(s)$ via Eq.~\eqref{eq:confidence} and flag skills with stronger evidence for review
        \STATE Reflexion on flagged skills: \textbf{keep} or \textbf{revise}
    \ENDIF
\ENDFOR
\RETURN $\pi_\theta, \mathcal{B}$
\end{algorithmic}
\end{algorithm}

\subsection{Formal Skill Definition and Representation}
\label{app:skill-definition}
In \textsc{SkillForge}, a skill is a callable, environment-verifiable decision unit that encodes reusable knowledge distilled from agent-environment interaction. Each skill is represented as a structured tuple:
\begin{align*}
  s = (&\, title,\; intent,\; principle, \\&\, applicability,\; category,\; status), \end{align*}
where $title$ is the unique callable identifier that the agent emits inside a \texttt{<skill\_call>} tag during rollout; $intent$ is a one-sentence description of the skill's purpose shown in the compact skill catalog; $principle$ is the core decision strategy returned to the agent upon invocation; $applicability$ specifies the trigger condition in terms of observable states or task properties; $category$ is either \texttt{general} or \texttt{specific}, used for retrieval routing; and $status$ is one of \texttt{active}, \texttt{under-review}, or \texttt{deprecated}, managed by the evidence-based verification mechanism.

Two properties distinguish skills from generic memory or prompt-injected advice. First, skills are callable: the agent sees only a compact catalog listing each skill's callable name and a one-line applicability trigger, and must explicitly emit a \texttt{<skill\_call>} tag to access the full content. Each invocation is recorded as a discrete event in the trajectory, making skill usage directly observable. Second, skills are environment-verifiable: because each invocation is linked to the final task outcome, per-skill statistics such as success rate and usage count can be computed from rollout data without additional annotation, and these statistics drive the confidence score and determine whether a skill should be revised or retained. At inference time, the agent interacts with skills in two stages: the catalog exposes up to $k_g + k_s$ entries containing only callable names and compact applicability triggers, and a successful skill call returns the corresponding intent, principle, and full applicability conditions in the next observation, keeping the prompt compact while preserving on-demand access to detailed guidance.

\subsection{Detailed Comparison with SkillRL}
  \label{app:skillrl-comparison}

  SkillRL is the closest prior work and also maintains a skill library that co-evolves with the agent's policy during
  reinforcement learning. Here we provide a more detailed comparison from both conceptual and empirical perspectives.

  Table~\ref{tab:skillrl-compare} summarizes the key conceptual differences. SkillRL injects retrieved skills as prompt context
  and uses trajectory-level success or failure to recursively expand the skill library. SkillForge instead lets the policy
  explicitly emit \texttt{<skill\_call>} tags during rollout, so that each invocation is recorded as a discrete event linked to the task
  outcome. This design enables skill-level credit assignment: the framework can identify which skill was actually used, whether
  it contributed to the outcome, and which specific skill should be revised after failure. SkillRL's recursive library growth
  can improve the skill bank over time, but the rollout does not reveal which retrieved skill was relied upon, making
  fine-grained verification difficult. In addition, SkillRL requires an SFT stage before RL training, whereas SkillForge
  operates in an RL-only setting without SFT initialization.

\begin{table}[h]
  \centering
  \setlength{\tabcolsep}{6pt}
  \begin{tabular}{@{}ll@{}}
  \toprule
  \textbf{Aspect} & \textbf{SkillRL / SkillForge} \\
  \midrule
  Pipeline & SFT + RL / RL only \\
  Skill use & Prompt injection / Explicit call \\
  Signal & Trajectory-level / Skill-level \\
  Update & Library expansion / Per-skill verify \\
  Optimization & Skills as context / Action + skill \\
  \bottomrule
  \end{tabular}
  \caption{Conceptual comparison between SkillRL and SkillForge.}
  \label{tab:skillrl-compare}
  \end{table}
  Empirically, under the same Qwen2.5-7B-Instruct backbone, SkillForge improves over SkillRL by +3.7 on ALFWorld, +10.3 on
  WebShop success rate, and raises AppWorld TGC/SGC from 19.0/5.36 to 23.8/14.3. Ablation results in the main paper further
  support this gap: removing explicit skill calling, effectiveness tracking, or LLM reflexion all degrades performance,
  confirming that the gains come from the proposed verification loop rather than superficial differences in setup.

  Table~\ref{tab:skillrl-efficiency} reports end-to-end training time on the same hardware. Despite adding explicit skill
  invocation, per-skill tracking, and teacher-driven verification, SkillForge uses less wall-clock time than SkillRL on all
  three benchmarks. This result is consistent with reusable skills reducing redundant exploration and shortening rollout episodes.

  \begin{table}[h]
  \centering
  \small
  \begin{tabular}{lcc}
  \toprule
  Benchmark & SkillForge & SkillRL \\
  \midrule
  AppWorld & 16.5h & 18.7h \\
  ALFWorld & 5.9h & 6.7h \\
  WebShop & 7.0h & 7.7h \\
  \bottomrule
  \end{tabular}
  \caption{End-to-end training time comparison.}
  \label{tab:skillrl-efficiency}
  \end{table}

\subsection{Teacher Sensitivity Analysis}
  \label{app:teacher-sensitivity}

  A natural question is how much of SkillForge's gain comes from the framework itself versus the capability of the external
  teacher LLM used for skill induction and revision. To isolate this factor, we fix the policy model as Qwen3-4B-Instruct and
  vary only the teacher model across three settings: no teacher (vanilla GRPO baseline), the policy model itself as teacher
  (self-teacher), and Qwen3-Max as teacher.

  \begin{table}[h]
  \centering
  \setlength{\tabcolsep}{6pt}
  \begin{tabular}{@{}ll@{}}
  \toprule
  \textbf{Setting} & \textbf{Score} \\
  \midrule
  \multicolumn{2}{@{}l}{\textit{WebShop}} \\
  GRPO (no teacher) & 79.4 \\
  Self-teacher (4B) & 82.0 \\
  Qwen3-Max teacher & 84.0 \\
  \midrule
  \multicolumn{2}{@{}l}{\textit{ALFWorld}} \\
  GRPO (no teacher) & 79.3 \\
  Self-teacher (4B) & 85.0 \\
  Qwen3-Max teacher & 87.9 \\
  \midrule
  \multicolumn{2}{@{}l}{\textit{AppWorld}} \\
  GRPO (no teacher) & 34.5 \\
  Self-teacher (4B) & 39.3 \\
  Qwen3-Max teacher & 44.6 \\
  \bottomrule
  \end{tabular}
  \caption{Teacher sensitivity analysis with Qwen3-4B-Instruct as the fixed policy model.}
  \label{tab:teacher-sensitivity}
  \end{table}

  Two observations follow. First, SkillForge improves over the no-teacher GRPO baseline on all three benchmarks even in the
  self-teacher setting, where the teacher is the same model as the policy. Since no stronger external model is involved, this
  gain cannot be attributed to teacher capability alone and suggests that the framework design itself contributes meaningfully.
  Second, replacing the self-teacher with Qwen3-Max further raises performance on all benchmarks, indicating that framework
  design and teacher quality are complementary: a stronger teacher produces higher-quality skill inductions and revisions, which
  the verification loop can then validate through environment interaction. Together, these results suggest that SkillForge's
  gains stem from both the explicit skill calling and verification mechanisms and the quality of the teacher, with neither
  factor alone fully accounting for the improvements.

\subsection{Efficiency Analysis}
  \label{app:efficiency}

  To quantify the computational overhead introduced by SkillForge, we decompose the skill-related training cost into two parts:
  (i) skill retrieval and calling ratio, measuring the time spent on skill retrieval and explicit skill calling during rollout,
  and (ii) skill update ratio, measuring the time spent on trajectory abstraction and dynamic skill generation or revision.

  \begin{table}[h]
  \centering
  \setlength{\tabcolsep}{6pt}
  \begin{tabular}{@{}ll@{}}
  \toprule
  \textbf{Metric} & \textbf{Value} \\
  \midrule
  \multicolumn{2}{@{}l}{\textit{AppWorld}} \\
  GRPO score / time & 34.5 / 16.1h \\
  SkillForge score / time & 44.6 / 16.5h \\
  Skill retrieve \& calling ratio & 2.48\% \\
  Skill update ratio & 1.70\% \\
  Total skill-related ratio & 4.18\% \\
  \midrule
  \multicolumn{2}{@{}l}{\textit{ALFWorld}} \\
  GRPO score / time & 79.3 / 6.9h \\
  SkillForge score / time & 87.9 / 5.9h \\
  Skill retrieve \& calling ratio & 6.56\% \\
  Skill update ratio & 2.92\% \\
  Total skill-related ratio & 9.48\% \\
  \midrule
  \multicolumn{2}{@{}l}{\textit{WebShop}} \\
  GRPO score / time & 79.4 / 7.2h \\
  SkillForge score / time & 84.0 / 7.0h \\
  Skill retrieve \& calling ratio & 7.22\% \\
  Skill update ratio & 2.25\% \\
  Total skill-related ratio & 9.47\% \\
  \bottomrule
  \end{tabular}
  \caption{Efficiency breakdown of SkillForge-specific overhead across three benchmarks.}
  \label{tab:efficiency}
  \end{table}

  The total skill-related overhead stays below 10\% across all three benchmarks. In terms of wall-clock time, SkillForge is
  faster than the GRPO baseline on ALFWorld and WebShop, and adds only 0.4h on AppWorld while improving the score from 34.5 to
  44.6. This suggests that reusable skills reduce redundant exploration and shorten rollout episodes, offsetting the cost of
  skill retrieval, calling, and update. The overhead is therefore modest relative to the consistent performance gains.

\subsection{Case Study: Skill Calling in Action}
\label{sec:appendix:case_study}

\begin{figure*}[b]
\centering
\includegraphics[width=\textwidth]{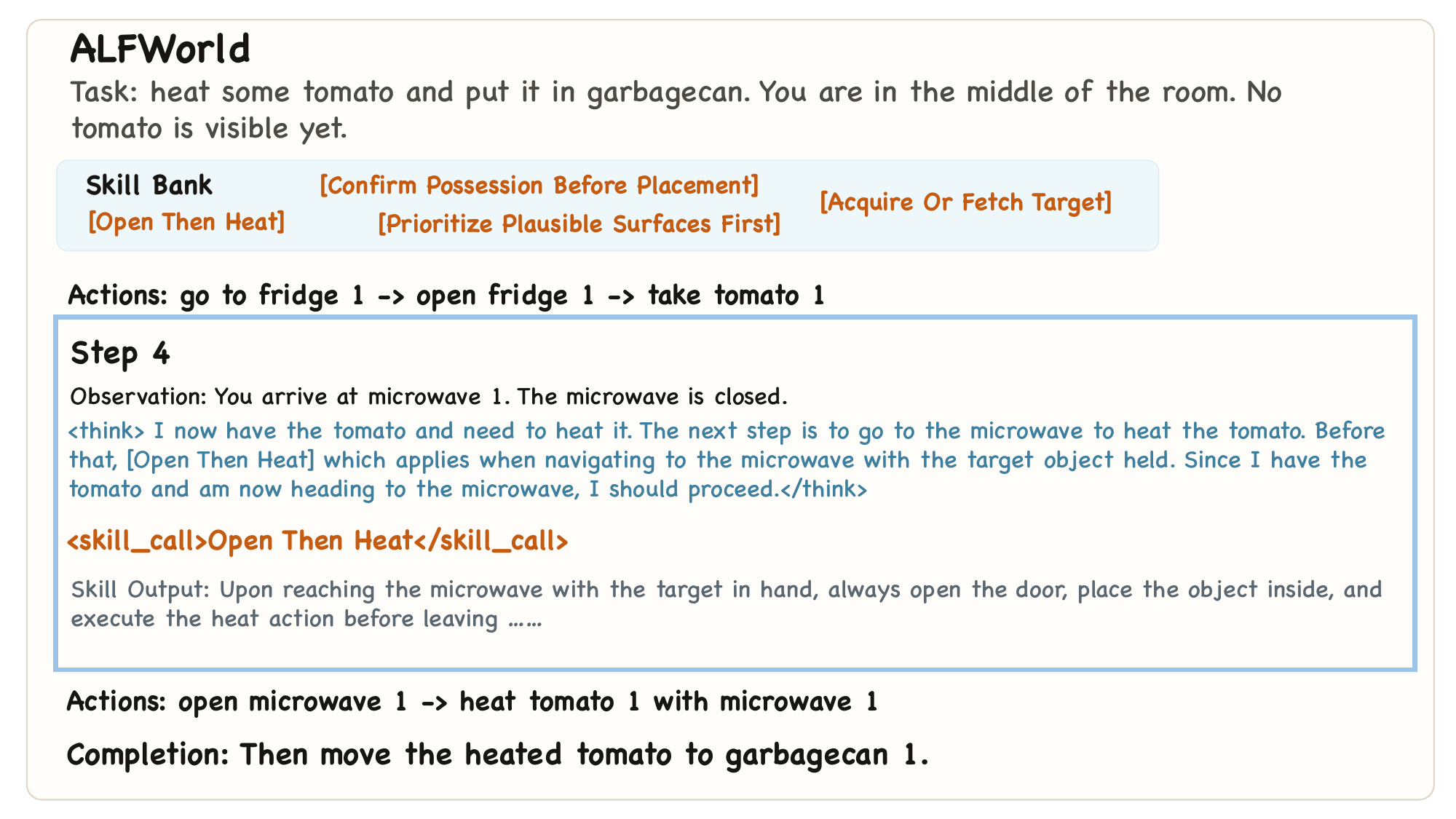}
\caption{Case study on ALFWorld. The agent invokes \emph{Open Then Heat} upon navigating to the microwave with the tomato in hand. The skill output provides the procedural sequence (open $\rightarrow$ heat), which the agent follows to complete the task successfully.}
\label{fig:case_alfworld}
\end{figure*}

\begin{figure*}[t]
\centering
\includegraphics[width=\textwidth]{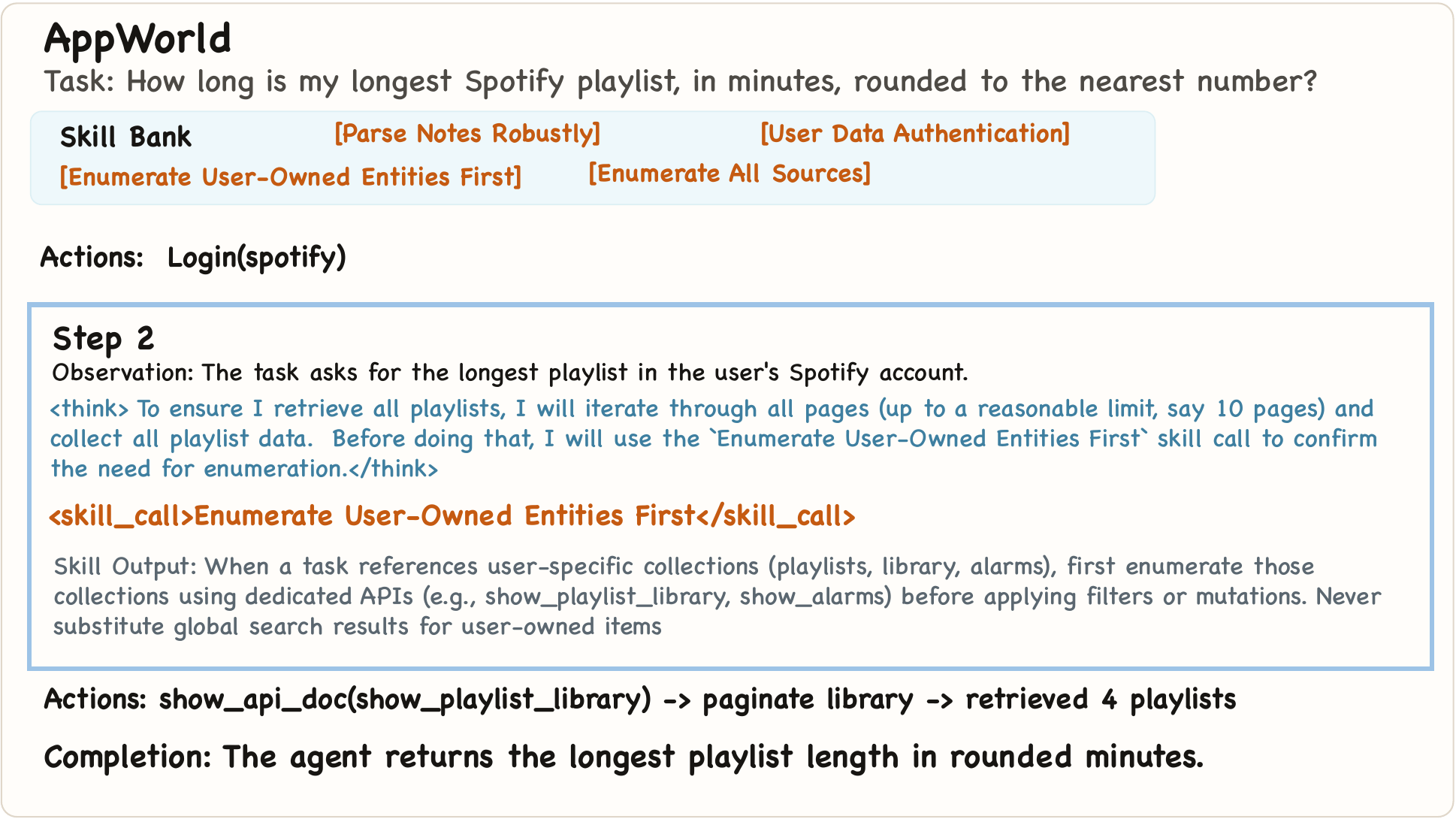}
\caption{Case study on AppWorld. The agent invokes \emph{Enumerate User-Owned Entities First} to scope playlist retrieval to user-owned data via \texttt{show\_playlist\_library} with pagination, avoiding the failure mode of substituting global search results for user-owned items.}
\label{fig:case_appworld}
\end{figure*}

\begin{figure*}[t]
\centering
\includegraphics[width=\textwidth]{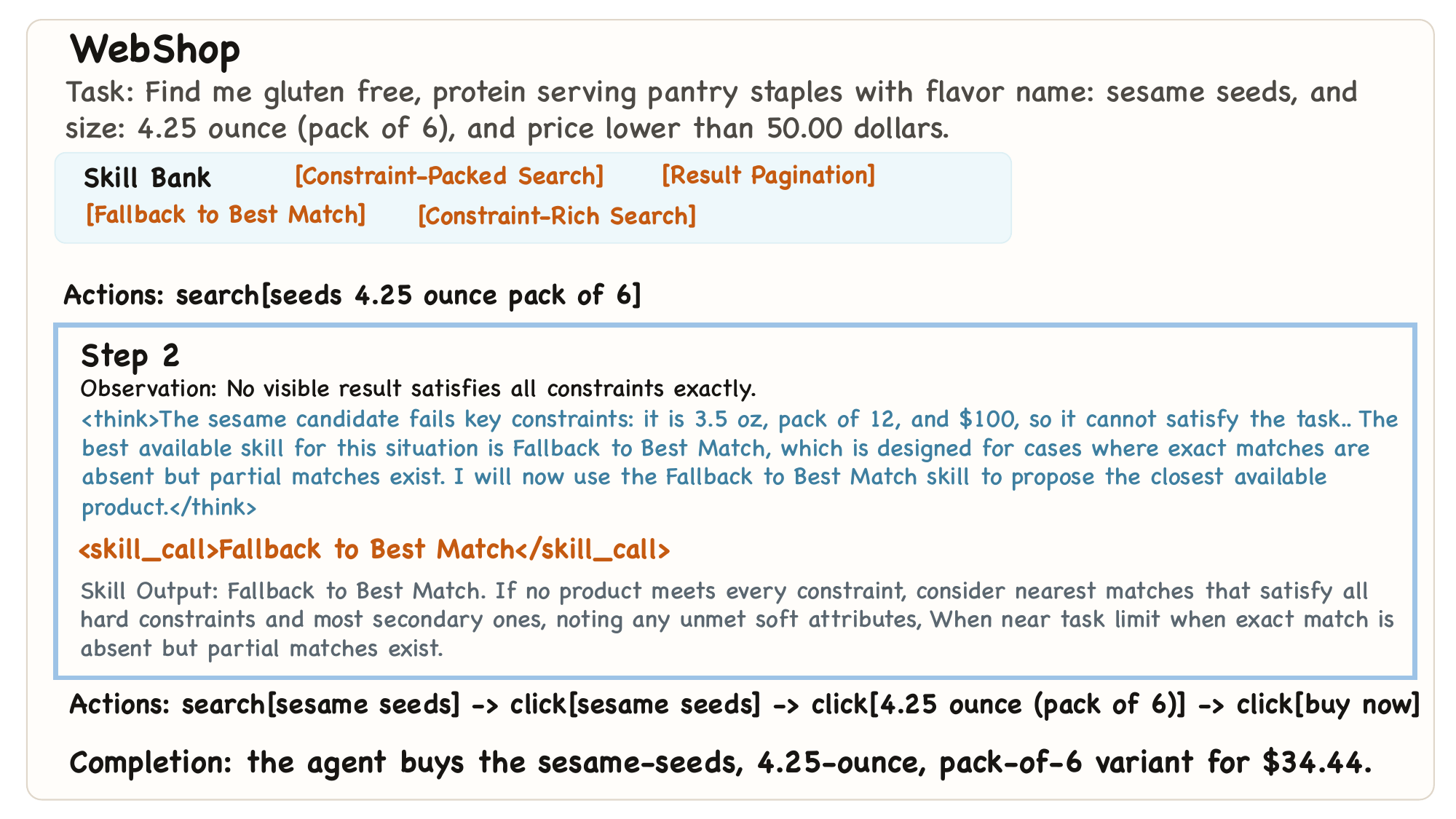}
\caption{Case study on WebShop. When no search result satisfies all constraints, the agent invokes \emph{Fallback to Best Match}. Guided by the skill, it clicks into the nearest candidate's product page, where hidden variant options (flavor: ``sesame seeds''; size: ``4.25 ounce (pack of 6)'') are progressively disclosed, enabling a successful purchase at \$34.44.}
\label{fig:case_webshop}
\end{figure*}

To illustrate how the agent leverages evolved skills during inference, we present representative rollout trajectories from three environments: ALFWorld (Fig.~\ref{fig:case_alfworld}), AppWorld (Fig.~\ref{fig:case_appworld}), and WebShop (Fig.~\ref{fig:case_webshop}). Each case highlights a clear cause-and-effect chain between skill invocation and subsequent agent behavior.

\paragraph{ALFWorld (Fig.~\ref{fig:case_alfworld}).}
The task is to heat a tomato and place it in the garbage can. No tomato is visible from the room center, so the agent first navigates to \texttt{fridge~1}, opens it to reveal \texttt{tomato~1} and \texttt{tomato~2}, and picks up \texttt{tomato~1}. With the target object now in inventory, the agent moves toward the microwave and simultaneously triggers \texttt{<skill\_call>Open Then Heat</skill\_call>}. The skill output returns the principle: ``Upon reaching the microwave with the target in hand, always open the door, place the object inside, and execute the heat action before leaving.'' Guided by this procedural reminder, the agent executes \texttt{open microwave~1} followed by \texttt{heat tomato~1 with microwave~1}, and then carries the heated tomato to \texttt{garbagecan~1} to complete the task. The skill transforms a generic navigation decision into a procedure-aware appliance interaction, preventing a common failure mode where agents attempt to heat the object without first opening the microwave.

\paragraph{AppWorld (Fig.~\ref{fig:case_appworld}).}
The task is to find the duration of the user's longest Spotify playlist in minutes. After authenticating and inspecting the API documentation, the agent recognizes that the task is scoped to user-owned data and invokes \texttt{<skill\_call>Enumerate User-Owned Entities First</skill\_call>}. The returned principle instructs the agent to enumerate user-specific collections via dedicated APIs (e.g., \texttt{show\_playlist\_library}) rather than substituting global search results. Following this guidance, the agent paginates through \texttt{show\_playlist\_library} and retrieves all 4 playlists, keeping the solution grounded in the user's own collection.

\paragraph{WebShop (Fig.~\ref{fig:case_webshop}).}
The task requires finding gluten-free pantry staples with flavor ``sesame seeds,'' size ``4.25 ounce (pack of 6),'' and price under \$50. The initial search returns no listing satisfying all constraints—the obvious sesame candidate fails on size (3.5~oz), pack count (12), and price (\$100). The agent triggers \texttt{<skill\_call>Fallback to Best Match</skill\_call>}, which advises considering nearest matches that satisfy all hard constraints. Guided by this strategy, the agent clicks into a near-match product page (Blue Diamond Nut Thins, \$34.44), where the environment \emph{progressively discloses} hidden variant options—selectable flavors including ``sesame seeds'' and sizes including ``4.25 ounce (pack of 6)''—that were not visible in the search listing. The agent selects both variants and completes the purchase at \$34.44.

Across all three environments, the case studies share a consistent pattern: the agent uses \texttt{<think>} reasoning to assess the current state, identifies a relevant skill from the catalog, invokes it via the \texttt{<skill\_call>} mechanism, and integrates the returned guidance into its subsequent actions. Notably, skills serve different roles depending on the environment: procedural sequencing in ALFWorld, data-access scoping in AppWorld, and adaptive search strategy in WebShop.

\subsection{Prompts Used in \textsc{SkillForge}}
\label{sec:appendix:prompts}

\begin{figure*}[t]
    \centering

    \begin{templatebox}{(a) Skill Catalog Injection (appended to first user message)}
## Skill (Optional)
You may append <skill_call>CALL_NAME</skill_call> after your normal action.
Use CALL_NAME exactly as shown in the catalog.
Do not break the required action format of the environment (e.g., keep <action>...</action> intact).

## Skill Catalog (General)
- call_name: {general_skill_1_call_name}
  description: {general_skill_1_trigger}
  ...

## Skill Catalog (Specific)
- call_name: {specific_skill_1_call_name}
  description: {specific_skill_1_trigger}
  ...
    \end{templatebox}

    \vspace{0.3em}

    \begin{templatebox}{(b) Skill Output (after successful call)}
## Skill Output
call_name: {matched_call_name}
title: {skill_title}
intent: {skill_intent}
principle: {skill_principle}
when_to_apply: {skill_trigger}
    \end{templatebox}

    \caption{Skill calling prompts: (a)~the calling instruction and compact catalog jointly appended to the first user message, listing only callable names and one-line trigger descriptions (up to $k_g{+}k_s$ entries), and (b)~the full skill content returned in the next observation after a successful \texttt{<skill\_call>} invocation.}
    \label{fig:template_skill_calling}
\end{figure*}

\begin{figure*}[t]
    \centering
    \begin{templatebox}{Prompt Template for Trajectory Abstraction}
You are an expert trajectory abstraction engine for an RL agent skill system. Abstract the raw trajectory into a reusable summary.

Task: {task}
Task Type: {task_type}
Domain: {domain}
Outcome: {outcome}
Domain-specific abstraction guidance: {domain_rules}

Return a JSON object with exactly these keys:
- contextual_description: short paragraph (1-3 sentences)
- refined_trajectory: chronological list of key steps; each item must contain step_index, action, critical_observation, reasoning
- strategic_guidelines: object with planning_pattern and mistakes_to_avoid
  * if SUCCESS: planning_pattern should be a reusable action chain string, mistakes_to_avoid should be []
  * if FAILURE: planning_pattern should be null, mistakes_to_avoid should contain abstract trigger_condition/bad_action pairs

Keep only the strategically critical steps; do not copy the whole trajectory.
Return JSON only.

Raw Trajectory:
{trajectory_text}
    \end{templatebox}
    \caption{Prompt for trajectory abstraction (\S\ref{sec:skill_calling}). Each rollout trajectory is compressed into a structured summary containing a contextual description, a refined key-step trajectory, and strategic guidelines. The resulting abstractions serve as input to the skill induction prompt (Fig.~\ref{fig:template_skill_induction}).}
    \label{fig:template_traj_abstraction}
\end{figure*}

\begin{figure*}[t]
    \centering
    \begin{templatebox}{Prompt Template for Multi-Pathway Skill Induction}
You are improving an RL agent's reusable skill bank from trajectory traces.
Harvest NEW skills that can be explicitly called in future trajectories.

Task-type routing:
- Dominant task_type: {dominant_type} | Domain guidance: {domain_instruction}
- Failed task_type counts: {failure_counts}
- Successful task_type counts: {success_counts}

Synthesis rules:
- Abstract reusable decision patterns from successful trajectories.
- Use failures only to sharpen when_to_apply boundaries.
- Compare patterns at category/task_type level, not same-task instance.

Constraints:
- Return a JSON array of 1 to {max_new_skills} skills.
- Required fields: skill_id, title (3-8 words), principle,
  when_to_apply (explicit trigger). Optional: intent, category, evidence.
- Avoid generic advice. Encode reusable strategy patterns.
- Do not duplicate existing skills by title/intent.

SUCCESSFUL TRAJECTORIES: {success_examples}
FAILED TRAJECTORIES: {failure_examples}
EXISTING SKILLS (dedup reference): {existing_skills}

Output format for each skill:
[{"skill_id":"{skill_id}", "title":"{skill_title}", "principle":"{skill_principle}", "when_to_apply":"{skill_trigger}", "category":"{category}"}]
    \end{templatebox}
    \caption{Prompt for multi-pathway skill induction (\S\ref{sec:skill_evolution}). The teacher LLM $M_T$ synthesizes new skills from successful and failed trajectory abstractions, with domain-specific routing and deduplication against existing skills.}
    \label{fig:template_skill_induction}
\end{figure*}

\begin{figure*}[t]
    \centering
    \begin{templatebox}{(a) Prompt for Skill-Aware Trajectory Abstraction}
You are analyzing an RL agent's trajectory to assess how a specific skill was used.

SKILL UNDER REVIEW:
  skill_id: {skill_id}
  title: {skill_title}
  principle: {skill_principle}
  when_to_apply: {skill_trigger}

TASK: {task_query}
OUTCOME: {SUCCESS or FAILURE}
TRAJECTORY (skill calls marked with SKILL_CALL): {trajectory_text}

Return a JSON object:
- contextual_description: 1-2 sentence task summary
- refined_trajectory: list of key steps (<= 80 chars each), mark skill call with [SKILL_CALL]
- skill_relevance: one of "causal" (skill directly caused success/failure), "incidental" (skill was used but outcome was unrelated), "harmful" (skill led to worse outcome)
- relevance_reasoning: 1 sentence explanation
    \end{templatebox}

    \vspace{0.8em}

    \begin{templatebox}{(b) Prompt for Skill Reflexion}
You are reviewing the effectiveness of a skill used by an RL agent.

SKILL:
  skill_id: {skill_id} | title: {skill_title}
  principle: {skill_principle} | when_to_apply: {skill_trigger}
  total_uses: {total_uses} | success_uses: {success_uses} | success_rate: {success_rate}
  consecutive_failures: {consec_fail}

USAGE CONTEXTS:
Context 1 [{outcome_1}]: Task: {task_query_1} | Summary: {contextual_description_1} | Skill relevance: {relevance_1}
...
The skill has been used {total_uses} times in total with {success_uses} successes
({overall_success_rate:.1\%} overall). Consider this full history when making your verdict.
A skill with good overall success rate should not be deprecated just because of recent failures.

Based on the skill definition and its usage contexts, choose ONE verdict:
- keep: skill is effective; no changes needed
- revise: principle or when_to_apply should be improved

Return a JSON object with these keys:
- verdict: one of "keep", "revise" | - reasoning: 1-2 sentence explanation
- revised_principle: (only if verdict=revise) improved principle text
- revised_when_to_apply: (only if verdict=revise) improved trigger condition
  \end{templatebox}

    \caption{Skill verification prompts (\S\ref{sec:skill_evolution}): (a)~trajectory abstraction with causal attribution (\textit{causal}/\textit{incidental}/\textit{harmful}), and (b)~reflexion verdict based on effectiveness statistics and abstracted usage contexts. The LLM returns \textit{keep} or \textit{revise} (with updated principle and trigger).}
    \label{fig:template_skill_reflexion}
\end{figure*}

We describe the full prompt templates used in the skill calling, induction, and verification stages of \textsc{SkillForge}. Fig.~\ref{fig:template_skill_calling} illustrates the skill calling procedure: (a)~the calling instruction and a compact skill catalog are appended to the first user message, and (b)~the full skill content is returned upon a successful \texttt{<skill\_call>} invocation. Fig.~\ref{fig:template_traj_abstraction} shows the trajectory abstraction prompt, which compresses each rollout into a structured summary (context description, refined key steps, and strategic guidelines) that serves as input to skill induction. Fig.~\ref{fig:template_skill_induction} presents the skill induction prompt, where the expert LLM synthesizes new skills from successful and failed trajectory abstractions via extraction, refinement, or contrastive pathways, while deduplicating against existing skills. Finally, Fig.~\ref{fig:template_skill_reflexion} depicts skill verification: (a)~trajectory abstraction with causal attribution, and (b)~the reflexion prompt where the LLM checks a flagged skill's definition against its usage contexts and outputs \textit{keep} or \textit{revise}.

\subsection{The Large Language Model Usage.}
During manuscript preparation, we use large language models (LLMs) to (i) improve grammar and spelling without altering the intended scientific content, and (ii) provide lightweight coding assistance (e.g., scripts and formatting help). All reported numerical results, analyses, and claims are produced by the authors. The authors design the methods, conduct the experiments, and verify the findings.

\end{document}